\pdfoutput=1
\documentclass{article}
\usepackage{iclr2027_conference,times}

\usepackage{amsmath,amsfonts,bm}

\def\eqref#1{equation~\ref{#1}}

\def\1{\bm{1}}

\DeclareMathAlphabet{\mathsfit}{\encodingdefault}{\sfdefault}{m}{sl}
\SetMathAlphabet{\mathsfit}{bold}{\encodingdefault}{\sfdefault}{bx}{n}

\usepackage[utf8]{inputenc}
\usepackage[T1]{fontenc}
\usepackage{hyperref}
\hypersetup{hidelinks,
  pdftitle={OPSRD: On-Policy Self-Role Distillation},
  pdfauthor={Weijie Ren, Yanwen Zhang, Hao Li, Zhuolin Qi, Hengyi Zhang, Naibo Wang}}
\usepackage{url}
\usepackage{graphicx}
\usepackage{booktabs}
\usepackage{multirow}
\usepackage{makecell}
\usepackage{xcolor}
\usepackage{colortbl}
\usepackage{amsmath,amssymb,mathtools}
\usepackage{microtype}
\usepackage{enumitem}
\usepackage{tabularx}

\definecolor{methodblue}{HTML}{DDEFFC}
\definecolor{softgray}{HTML}{F1F4F7}
\definecolor{inkblue}{HTML}{285781}
\newcommand{\method}{\textsc{OPSRD}}
\newcommand{\avg}{\mathrm{Avg@12}}
\newcommand{\pp}{\,\mathrm{pp}}

\title{OPSRD: On-Policy Self-Role Distillation}

\author{Weijie Ren$^{1*}$\enspace Yanwen Zhang$^{2*}$\enspace Hao Li$^{3*}$\enspace Zhuolin Qi$^{1}$\enspace Hengyi Zhang$^{1}$\enspace Naibo Wang$^{1\dagger}$ \\
\textnormal{$^1$Zhejiang University\quad $^2$University of Electronic Science and Technology of China} \\
\textnormal{$^3$University of Science and Technology of China} \\
\textnormal{\small\texttt{3200101501@zju.edu.cn}, \texttt{2023091601016@std.uestc.edu.cn},} \\
\textnormal{\small\texttt{haoli2101@mail.ustc.edu.cn}, \texttt{qizhuolin666@gmail.com},} \\
\textnormal{\small\texttt{22651274@zju.edu.cn}, \texttt{wangnaibo@zju.edu.cn}} \\
\textnormal{\small $^*$Equal contribution.\quad $^\dagger$Corresponding author.}}

\iclrfinalcopy

\begin{document}
\maketitle
\lhead{Preprint}

\begin{abstract}
Role prompting elicits specialized behavior from large language models through an expert identity, offering a lightweight way to guide reasoning on demanding tasks. However, evaluating or distilling complete role-prompted answers can miss useful next-token preferences when the sampled solution remains incorrect. Transferring these preferences also requires an objective that reaches alternatives the student rarely predicts. We introduce \method{}, which uses a fixed expert role as privileged teaching context for on-policy self-distillation without reference solutions. A role-free student generates a trajectory, and a frozen instance of the same base model supplies role-conditioned distributions on its exact prefixes, exposing alternatives beyond the sampled continuation. Teacher-weighted forward KL targets alternatives the student underestimates, with clipping to limit individual vocabulary contributions. Supervision is restricted to the highest-entropy half of student positions, concentrating learning where predictions are uncertain. Experiments on three competition-math benchmarks with Qwen3-1.7B, 4B, and 8B show improvements over the base models without role prompts at inference. Forward KL achieves the highest macro-averaged accuracy among the three evaluated divergences \mbox{at every scale}. Code is available at \url{https://github.com/zhansan114514/OPSRD}.
\end{abstract}

\section{Introduction}
\label{sec:introduction}

Large language models serve as general-purpose assistants for reasoning, explanation, and problem solving. Their usefulness depends on how instructions elicit the capabilities needed for a particular task \citep{liu2023promptingsurvey}. Role prompting makes this dependence explicit by asking a model to adopt an expert identity \citep{kong2023expertprompting}. A mathematical expert role can request careful derivations and verification, while a critic role can encourage checking intermediate conclusions. Such prompts offer a practical way to guide model behavior without changing its parameters.

However, existing uses of role prompting often assess expertise through complete generated responses. Reported reasoning gains and failures \citep{kong2024betterroleplay,kim2024persona} therefore mix the role's local influence with the effects of sampling an entire solution. A useful next-token preference can remain unsampled or be followed by a later mistake. Distilling expert-prompted responses \citep{kong2023expertprompting} likewise transfers only the sampled continuations. On-policy self-distillation accesses richer token distributions, but solution-conditioned OPSD derives its teaching advantage from a reference solution for each problem \citep{zhao2026opsd}. The open question is whether an expert role can supervise the student's reasoning even when its complete answers offer \mbox{no measured advantage}.

Our starting observation is a negative result. On three competition-math benchmarks, adding an expert olympiad mathematician role to Qwen3-1.7B changes macro accuracy from 35.74 to 34.54, providing no measured improvement over the ordinary prompt. A neutral system instruction reaches the same score. The expert label alone is insufficient to improve sampled solutions in this setting. What remains unobserved is the role-conditioned distribution at each intermediate choice, before later decisions determine the final answer.

We hypothesize that a role-conditioned model can teach useful next-token preferences even when its complete solutions offer no accuracy gain. Under this hypothesis, the role's value lies in shifting probability toward helpful continuations at uncertain student prefixes. Holding these prefixes fixed lets the teacher evaluate the choices the student actually faces. On-policy distillation provides this alignment \citep{agarwal2024onpolicy,gu2024minillm}, allowing us to test a reusable role as privileged teaching context without per-problem reference solutions.

We develop \method{} to test this hypothesis. The student generates trajectories under its ordinary prompt. A frozen instance of the same base model receives the expert role and scores the exact student prefixes, supplying a distribution over possible next tokens. We retain the highest-entropy half of student positions to focus supervision on uncertain choices. Only a low-rank student adapter is updated \citep{hu2022lora}; the teacher stays fixed, no reference answer enters training, and the trained student uses its ordinary prompt at inference.

For this feedback to help, the objective must reach alternatives the student currently underestimates. A reverse-KL update weakens as the student's probability of an alternative approaches zero, even when the teacher supports it. Forward KL weights the target by teacher probability and directly corrects this underestimation before clipping. We therefore use forward KL, with a cap on individual vocabulary contributions to limit large discrepancies from an imperfect teacher. We compare forward KL, reverse KL, and Jensen--Shannon divergence under the same role and entropy mask.

Experiments cover Qwen3-1.7B, 4B, and 8B on three competition-math benchmarks. Forward KL achieves the highest macro accuracy among the evaluated divergences at all three scales (Figure~\ref{fig:results}). At 1.7B, \method{} improves Base by 6.30 percentage points in the primary run, and independent training repeats retain the gain. Placement controls favor keeping the role in the teacher. The study contributes an empirical contrast between direct prompting and role-based supervision, an answer-free on-policy method, and an analysis of how divergence and role placement govern this transfer.

\begin{figure*}[t]
  \centering
  \includegraphics[width=0.99\textwidth]{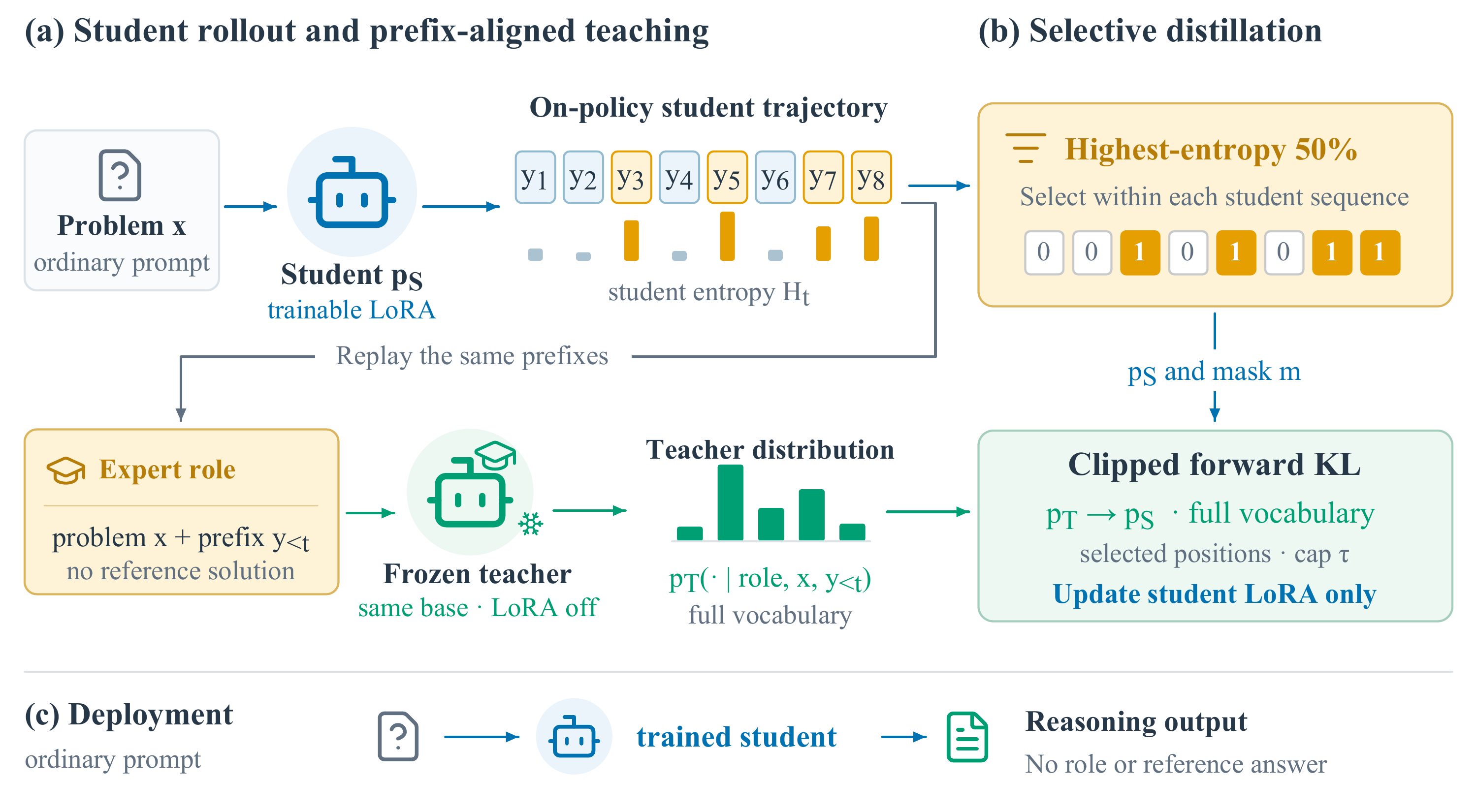}
  \caption{\method{} overview. The student generates a role-free trajectory; the frozen expert-role teacher evaluates the same prefixes. The highest-entropy half of student positions receive clipped, full-vocabulary forward KL. Only student LoRA parameters are updated, and deployment uses the ordinary prompt at test time.}
  \label{fig:method}
\end{figure*}

\section{Related Work}
\label{sec:related}

\subsection{On-Policy Distillation}

On-policy distillation learns from student-generated trajectories, with the teacher evaluating the same prefixes \citep{agarwal2024onpolicy,gu2024minillm}. Sequence-level distillation instead trains on teacher-generated outputs \citep{kim2016sequence}. Using student prefixes addresses the train--test distribution mismatch central to imitation learning \citep{ross2011dagger}. The teacher supplies distributional supervision \citep{hinton2015distilling}, while the student determines where it applies. Generalized knowledge distillation permits different divergences, and MiniLLM uses reverse KL. Entropy-aware OPD adds forward KL when teacher entropy is high \citep{jin2026entropyaware}, linking the objective \mbox{to teacher uncertainty}.

The quality of feedback also depends on the prefixes the student visits. Speculative knowledge distillation lets the teacher replace poorly ranked student proposals, bringing the sampled trajectory closer to teacher-supported continuations \citep{xu2025speculative}. Other methods adjust the learning signal at a fixed prefix. Asymmetric OPD combines positive reinforcement with local divergence minimization \citep{jia2026asymmetric}, while empirical analyses identify unreliable teacher guidance and imbalanced token supervision as failure modes \citep{fu2026revisiting}. CROP selects positions using counterfactual task relevance \citep{li2026crop}, and spurious-signal filtering uses input dependence together with divergence to remove misleading updates \citep{jiang2026mislead}. This work establishes that visiting student states alone does not make every teacher preference equally useful.

Self-distillation changes where the teacher's advantage comes from. Earlier approaches learn from model-generated reasoning, instructions, or filtered outputs \citep{zelikman2022star,wang2023selfinstruct,gulcehre2023rest}. OPSD uses one model under two contexts: its teacher sees a privileged solution and evaluates the question-only student's rollouts \citep{zhao2026opsd}. This supplies token-level feedback without a separately pretrained teacher. \method{} builds on this contextual asymmetry, replacing the per-problem solution with a reusable expert role and asking how to transfer that weaker source of supervision.

\subsection{Role-Conditioned Supervision}

Prompts can expose behavior that a model does not reliably express under an ordinary instruction. ExpertPrompting constructs an expert identity for each instruction, uses it to generate responses, and trains ExpertLLaMA on the resulting data \citep{kong2023expertprompting}. Role-play prompting also improves selected reasoning tasks \citep{kong2024betterroleplay}, and RoleLLM studies how to elicit and train broader role-playing abilities \citep{wang2024rolellm}. Controlled experiments find that persona prompts can impair reasoning as well as help it \citep{kim2024persona}. The question for distillation is consequently how a role's influence should be measured and transferred when its complete sampled answers offer an uncertain advantage.

Context distillation provides a bridge from conditional behavior to learned behavior by internalizing gains elicited by instructions and scratchpads \citep{snell2022context}. In-context learning distillation transfers the ability to learn from demonstrations \citep{huang2022incontext}, and self-distillation can reduce distribution gaps during fine-tuning \citep{yang2024distributiongap}. These studies show several ways that a model's context can create a teaching signal. For a role teacher, the relevant signal may lie in its preferences among possible next tokens on an existing student prefix. Capturing these preferences gives access to guidance that a comparison of complete role-prompted solutions can miss.

In \method{}, a single role supplies context for a teacher that shares the student's base model. The teacher scores role-free student prefixes, and entropy selection concentrates forward-KL supervision on uncertain positions. This makes the role's distributional feedback the object of distillation while keeping student generation under the ordinary prompt.

\section{Method}
\label{sec:method}

\subsection{Setting}

Let $x$ be a math problem and let $c_S$ denote the role-free training context. The student $p_\theta$ contains low-rank trainable parameters on top of a pretrained model $p_{\theta_0}$. For each problem, the student samples a reasoning trajectory
\begin{equation}
 y_{1:T} \sim p_\theta(\cdot\mid c_S,x),
 \label{eq:rollout}
\end{equation}
without a role or reference solution. Sampled tokens are detached. The frozen teacher uses the same base weights with the student adapter disabled and an expert-role context $c_T$. Training uses a non-thinking student template and a thinking teacher template; evaluation enables thinking. Training also includes a \texttt{Problem:} prefix that evaluation omits. Appendix~\ref{app:prompt} specifies the complete prompts.

At position $t$, both models receive the exact student prefix $y_{<t}$. Let $z_S^t$ and $z_T^t$ be their logits. The distributions used in entropy selection and distillation apply temperature $\eta=1.1$,
\begin{align}
 p_S^t(v) &= \operatorname{softmax}(z_S^t/\eta)_v, \quad z_S^t=z_\theta(c_S,x,y_{<t}), \\
 p_T^t(v) &= \operatorname{softmax}(z_T^t/\eta)_v, \quad z_T^t=z_{\theta_0}(c_T,x,y_{<t}),
 \label{eq:views}
\end{align}
for every vocabulary item $v$. The teacher evaluates exactly the tokens sampled by the student. Sampling additionally applies top-$p=0.95$ and top-$k=20$; the loss distributions retain the full vocabulary. Only student logits receive gradients; sampling and teacher probabilities remain detached.

\subsection{Role-privileged teaching context}

\begin{figure*}[t]
  \centering
  \includegraphics[width=0.99\textwidth]{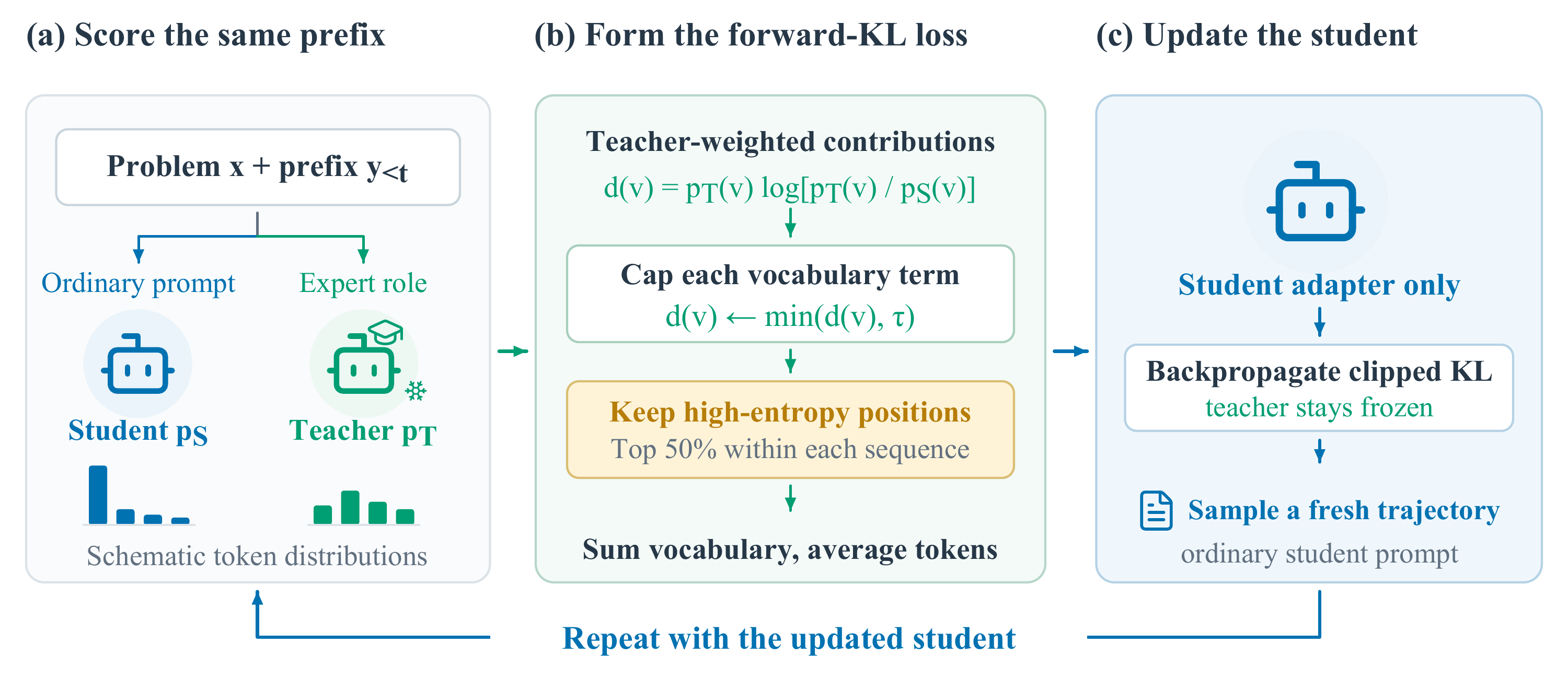}
  \caption{One \method{} update. Student and teacher score the same prefix; clipped vocabulary contributions are summed and averaged over the selected high-entropy positions. Only the student adapter is updated before the next rollout. The small distributions are schematic.}
  \label{fig:trainingflow}
\end{figure*}

The teacher's reusable role requests rigorous olympiad reasoning, verification, and a checked final answer, without task-specific hints. Holding $(x,y_{<t})$ fixed aligns the compared positions. Their distributional difference includes the role, thinking template, and adapter state. The neutral-teacher control shares the template and chat placement but uses a generic instruction. Figure~\ref{fig:method} shows the overall method; Figure~\ref{fig:trainingflow} follows one training update.

\subsection{Student-entropy selection}

The role does not need equal authority everywhere. Many tokens are nearly fixed by syntax or an earlier decision, so we compute student entropy at each valid position,
\begin{equation}
 H_t=-\sum_{v\in\mathcal{V}} p_S^t(v)\log p_S^t(v),
 \label{eq:entropy}
\end{equation}
and retain the $k=\max(1,\lceil\rho T\rceil)$ largest values within each sequence, with default $\rho=0.5$. This assigns every sequence its own selection quota $m_t\in\{0,1\}$. Longer sequences still contribute more selected positions to the token-averaged objective. Selection uses detached student logits, so the ranking receives no gradient; Appendix~\ref{app:algorithm} illustrates the selection rule.

\subsection{Clipped forward-KL objective}

For a selected position, the unmodified forward KL is
\begin{equation}
 D_t=\sum_{v\in\mathcal{V}}p_T^t(v)\log\frac{p_T^t(v)}{p_S^t(v)}.
 \label{eq:fkl}
\end{equation}
We retain the full vocabulary and cap each signed contribution at $\tau=0.05$ for the 1.7B and 4B runs and $\tau=0.06$ for the 8B runs. Writing $d_{t,v}=p_T^t(v)\log[p_T^t(v)/p_S^t(v)]$, the batch objective is
\begin{equation}
 \mathcal{L}(\theta)=
 \frac{\sum_{i\in\mathcal B}\sum_{t=1}^{T_i}m_{i,t}
 \sum_{v\in\mathcal{V}}\min\{d_{i,t,v},\tau\}}
 {\max(1,\sum_{i\in\mathcal B}\sum_{t=1}^{T_i}m_{i,t})}.
 \label{eq:loss}
\end{equation}
The clip limits a few large teacher preferences without truncating the vocabulary or changing the selected positions. The unmodified KL is nonnegative, but its vocabulary contributions have both signs. Capping only positive contributions can make the optimized sum negative. We therefore log the unclipped divergence separately from the training loss. Only LoRA parameters receive gradients; the teacher remains fixed.

\subsection{Why forward KL fits a role teacher}

The teacher's contribution is a distribution over alternatives at an existing student prefix. Forward KL gives each alternative weight proportional to its teacher probability. To see the resulting update, consider an unclipped selected position. Its derivative with respect to student logit $z_{S,j}$ is
\begin{equation}
 \frac{\partial D_{\mathrm{F}}}{\partial z_{S,j}}
 =\frac{p_S(j)-p_T(j)}{\eta}.
 \label{eq:fgrad}
\end{equation}
When the teacher assigns appreciable probability to a continuation the student rarely predicts, this gradient directly increases its logit. For reverse KL, $D_{\mathrm{R}}=\sum_v p_S(v)\log[p_S(v)/p_T(v)]$, the derivative instead becomes
\begin{equation}
 \frac{\partial D_{\mathrm{R}}}{\partial z_{S,j}}
 =\frac{p_S(j)}{\eta}\left(\log\frac{p_S(j)}{p_T(j)}-D_{\mathrm{R}}\right).
 \label{eq:rgrad}
\end{equation}
For fixed positive teacher probabilities, this update vanishes as $p_S(j)$ approaches zero. The distinction motivates forward KL for transferring preferences that the role elicits but the student underuses. It also complements entropy selection, which directs learning toward prefixes with several plausible continuations. The contribution cap modifies these derivatives when active; Appendix~\ref{app:algorithm} gives the exact clipped forward gradient. The following experiments evaluate this choice under the implemented loss and training procedure.

\subsection{Why the asymmetry matters}

Teacher-only conditioning keeps the student persona-free during both rollout and evaluation. Shared or student-only conditioning changes the prefixes on which the teacher supplies supervision. Appendix~\ref{app:algorithm} contrasts the three placements. Their comparison tests whether role-conditioned student sampling helps the final model under either evaluation prompt. Thinking-mode and formatting choices remain fixed across the placement ablation, as specified in Appendix~\ref{app:prompt}.

\section{Experiments}
\label{sec:experiments}

Our experiments test whether an expert role can provide useful supervision when it offers little benefit as a direct prompt. We first compare role distillation with prompting and answer-conditioned distillation, then examine the choices that govern the transfer. Repeated training and evaluation assess the stability of the gains, and a problem-level analysis shows how those gains change \mbox{the student's answers}.

\subsection{Experimental Setup}

\paragraph{Models and data.}
We use Qwen3-1.7B, 4B, and 8B \citep{yang2025qwen3}, with a frozen instance of each base model serving as its own teacher. Training uses the 29,434-example mathematical split derived from OpenThoughts \citep{guha2025openthoughts}. All methods train a rank-64 LoRA adapter for 100 optimizer steps with learning rate $5\times10^{-6}$ and effective batch size 36. The student samples one rollout per problem, capped at 1,024 tokens. We use 1.7B for the detailed ablations and independent training repeats; the larger models test whether the recipe remains useful as base accuracy increases. Appendix~\ref{app:repro} gives the remaining hyperparameters and exact prompts.

\paragraph{Evaluation.}
We evaluate on AIME 2024, AIME 2025, and HMMT February 2025, each containing 30 problems. Every problem receives 12 samples at temperature 1.0 and top-$p$ 0.95, with a 38,912-token generation cap. We report average sample accuracy for each task and its unweighted macro average, denoted $\avg$. All comparisons use the final step-100 checkpoint and the ordinary evaluation prompt unless a prompt change is the intervention being tested. Reported 95\% intervals use paired problem-level bootstrap resampling, with the training repeats additionally accounting for shared test problems. Appendix~\ref{app:statistics} defines these estimators.

\paragraph{Baselines.}
Base is the untrained model. Direct expert and neutral prompting test whether an extra system message improves inference by itself. Answer OPSD supplies a reference solution to the teacher, following \citet{zhao2026opsd}, and provides the main training baseline. Role full and Neutral full distill an expert or generic teacher instruction at every valid position. \method{} combines the expert teacher with the highest-entropy half of student positions. This set of comparisons connects the practical benefit of role distillation to the teacher context, role placement, and training objective.

\subsection{Main Results}

\begin{table*}[t]
\caption{Primary Qwen3-1.7B comparison. Avg@12 in percent; gains in percentage points.}
\label{tab:main}
\centering\small
\setlength{\tabcolsep}{4.5pt}
\begin{tabular}{lrrrrr}
\toprule
Method & AIME24 & AIME25 & HMMT25 & Macro & $\Delta$ Base \\
\midrule
Base & 46.67 & 36.11 & 24.44 & 35.74 & $0.00$ \\
Answer OPSD & 51.39 & 39.17 & 25.56 & 38.70 & $+2.96$ \\
Base with expert role & 46.94 & 37.22 & 19.44 & 34.54 & $-1.20$ \\
Base with neutral prompt & 45.83 & 35.83 & 21.94 & 34.54 & $-1.20$ \\
Role full & 55.56 & 39.72 & 25.00 & 40.09 & $+4.35$ \\
\rowcolor{methodblue}\method{} & \textbf{57.78} & \textbf{41.67} & \textbf{26.67} & \textbf{42.04} & $+6.30$ \\
Neutral full & 50.28 & 40.00 & 24.44 & 38.24 & $+2.50$ \\
\bottomrule
\end{tabular}
\end{table*}

\paragraph{From prompting to supervision.}
The central result is a change in how the same role becomes useful. Direct expert prompting gives 34.54 macro accuracy, close to the 35.74 Base score, and the paired difference has an interval of $[-4.35,1.94]$. Using that role for the teacher raises accuracy to 40.09 with full distillation and 42.04 with \method{} (Table~\ref{tab:main}). The latter improves Base by $6.30\pp$, with interval $[3.24,9.54]$, and Answer OPSD by $3.33\pp$, with interval $[0.83,6.02]$. The teacher receives the same short instruction for every training example, making the supervision reusable across problems without their reference solutions.

The benchmark results clarify this change. Direct prompting lowers HMMT accuracy while leaving the two AIME scores near Base. After distillation, all three task point estimates improve, with the largest gain on AIME 2024. The role's failure to raise aggregate prompting accuracy therefore does not preclude useful feedback on the student's prefixes. The placement and divergence ablations below examine how that feedback reaches the trained model.

\begin{table*}[t]
\caption{Macro Avg@12. The 4B summary averages four evaluation repeats.}
\label{tab:scale}
\centering\small
\setlength{\tabcolsep}{4.5pt}
\begin{tabular}{llrrrrr}
\toprule
Model & Evaluation & Base & Answer OPSD & \cellcolor{methodblue}\method{} & $\Delta$ Base & $\Delta$ Answer \\
\midrule
Qwen3-1.7B & Primary & 35.74 & 38.70 & \cellcolor{methodblue}42.04 & \cellcolor{methodblue}$+6.30$ & $+3.33$ \\
Qwen3-4B & Mean of 4 & 60.86 & 61.50 & \cellcolor{methodblue}62.52 & \cellcolor{methodblue}$+1.67$ & $+1.02$ \\
Qwen3-8B & Primary & 62.78 & 64.26 & \cellcolor{methodblue}65.83 & \cellcolor{methodblue}$+3.06$ & $+1.57$ \\
\bottomrule
\end{tabular}
\end{table*}

\paragraph{Results across model sizes.}
Table~\ref{tab:scale} extends the comparison to stronger base models. \method{} improves macro accuracy at all three scales, with gains of 6.30, 1.67, and 3.06 points over Base. The 4B entry averages four evaluations of each fixed checkpoint; the other scales use the primary evaluation. The gains are smaller as baseline accuracy rises, although their magnitude does not change monotonically with model size. Across these summaries, role distillation also exceeds the answer-conditioned comparator while requiring less privileged training context.

The per-task comparisons reveal different preferences for teacher context. Answer OPSD has the strongest 4B primary evaluation, and it slightly exceeds \method{} on 8B AIME 2025. Averaging the repeated 4B evaluations changes the overall ordering. Role distillation remains useful across model sizes, while the strength of its advantage depends on the benchmark and evaluation sample. Appendix~\ref{app:fullresults} provides the complete benchmark scores, and Section~\ref{sec:robustness} assesses variation across all four repeated evaluations.

\subsection{Ablation Studies}
\label{sec:ablations}

\paragraph{Teacher context.}
We first ask how much of the benefit requires an expert instruction. Neutral full improves Base by 2.50 points, showing that a generic teacher context already supplies useful self-distillation feedback. Role full adds 1.85 points over this control, with interval $[-0.28,3.98]$. The combined \method{} recipe is 3.80 points above Neutral full, with interval $[1.20,6.39]$. These results motivate the expert context, while the missing neutral-teacher 50\% control leaves its contribution at the selected positions unresolved. The remaining ablations hold the expert instruction fixed and test how to transfer its feedback.

\paragraph{Role placement.}
The role can change either the student's trajectory or the teacher's view of it. To distinguish these effects, we place the role in the teacher, the student, or both, holding the model size, selection fraction, checkpoint, and evaluation protocol fixed. Each trained model is then evaluated with and without the role. This comparison tests the division of labor in \method{}, where the student chooses the trajectory and the teacher supplies the additional perspective.

\begin{table}[t]
\caption{Role placement and deployment prompt on Qwen3-1.7B.}
\label{tab:locus}
\centering\small
\setlength{\tabcolsep}{4.5pt}
\begin{tabular}{llr}
\toprule
Training role & Test role & Macro \\
\midrule
\rowcolor{methodblue}Teacher only & No role & 42.04 \\
Teacher only & Expert role & 39.44 \\
Shared role & No role & 38.98 \\
Shared role & Expert role & 36.94 \\
Student only & No role & 38.24 \\
Student only & Expert role & 36.30 \\
\bottomrule
\end{tabular}
\end{table}

Teacher-only conditioning achieves the highest score under the ordinary evaluation prompt (Table~\ref{tab:locus}). Giving the role to both models lowers accuracy by 3.06 points, and giving it only to the student lowers accuracy by 3.80 points. Both paired intervals lie below zero. Adding the role again at evaluation lowers each checkpoint's point estimate, including the teacher-only model. The benefit of training with role feedback therefore persists when the student returns to its ordinary prompt. Conditioning the rollout generator on the role changes the training context in a way that these placement results \mbox{do not favor}.

\begin{figure*}[t]
  \centering
  \includegraphics[width=0.98\textwidth]{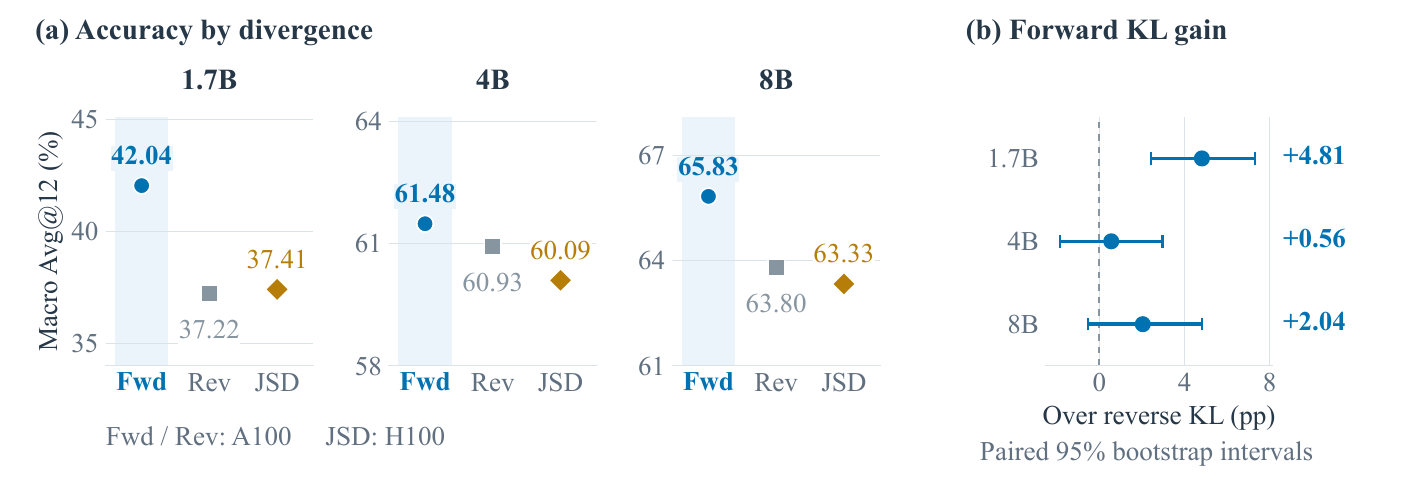}
  \caption{Divergence ablation. (a) Macro Avg@12 from the primary evaluation at each scale. (b) Forward-minus-reverse gains with paired 95\% problem-bootstrap intervals. Each condition uses one training run; JSD uses H100 and the KL runs use A100.}
  \label{fig:results}
\end{figure*}

\paragraph{Distillation objective.}
\label{sec:kl}
The next comparison tests how the student should absorb the role teacher's distribution. We replace forward KL with reverse KL or JSD while retaining the expert teacher and 50\% entropy mask. Forward KL gives the highest macro accuracy at all three scales (Figure~\ref{fig:results}). The clearest separation is at 1.7B, where it exceeds reverse KL by 4.81 points with interval $[2.41,7.31]$. This result agrees with the motivation in Section~\ref{sec:method}: teacher-supported alternatives receive a direct learning signal even when the student assigns them little probability.

The larger-model gaps are less conclusive. Forward-minus-reverse intervals cross zero at 4B and 8B, and reverse KL is stronger on 4B AIME 2024. We therefore use forward KL as the default supported most clearly by the primary scale. All divergence variants have one training run; JSD also differs in hardware and is supported by saved aggregate scores. Appendix~\ref{app:kl} gives the objective definitions, full task results, and comparison settings. The 4B result in this figure uses the primary evaluation, whereas Table~\ref{tab:scale} summarizes the repeated evaluations.

\paragraph{Supervision density.}
Finally, we test whether the gain survives concentrating supervision on uncertain positions. Selecting half the valid positions raises the primary score by 1.94 points over Role full, with interval $[-0.56,4.44]$. Across three independent training runs, full and selective distillation remain close, with a mean difference of 0.77 points and interval $[-1.33,2.90]$. The supported finding is that half the supervised positions retain the benefit of role distillation. Both variants still compute the student and teacher forward passes.

The entropy measurements help explain what this restriction does. Selected positions have mean student entropy 0.874, compared with 0.023 at omitted positions. The objective therefore focuses on predictions with substantial uncertainty while leaving nearly settled predictions outside the loss. A smaller sampling-budget pilot also explored 25\% and 75\% selection, but did not establish a better fraction. Appendix~\ref{app:entropy} reports this sensitivity study separately from the full-budget comparison.

\subsection{Robustness}
\label{sec:robustness}

\begin{table}[t]
\caption{Mean macro Avg@12 and sample standard deviation over three training runs at 1.7B and four evaluations of fixed checkpoints at 4B. The fixed 1.7B Base score is 35.74.}
\label{tab:robustness}
\centering\small
\setlength{\tabcolsep}{4.5pt}
\begin{tabular}{lllr}
\toprule
Model & Repeat & Method & Macro \\
\midrule
1.7B & Training & Role full & $39.48\pm1.32$ \\
\rowcolor{methodblue}1.7B & Training & \method{} & $40.25\pm1.62$ \\
4B & Evaluation & Base & $60.86\pm0.67$ \\
4B & Evaluation & Answer OPSD & $61.50\pm0.77$ \\
\rowcolor{methodblue}4B & Evaluation & \method{} & $62.52\pm0.78$ \\
\bottomrule
\end{tabular}
\end{table}

We assess whether the gains survive changes in optimization and evaluation sampling. At 1.7B, three independent training runs give \method{} a mean accuracy of $40.25\pm1.62$, improving the fixed Base evaluation by 4.51 points with interval $[1.54,7.69]$. Every run improves Base. Role full also retains its gain, consistent with the supervision-density ablation. The repeat results support the value of role distillation more strongly than a further advantage from selecting half the positions.

At 4B, four evaluations of each fixed checkpoint yield a mean gain of $1.67\pp$ over Base, with sample standard deviation $0.39\pp$ for the paired gains. \method{} exceeds Base in every evaluation. This agreement is useful because the primary comparison with Answer OPSD changes when evaluation sampling is repeated. Table~\ref{tab:robustness} summarizes all repeats. These evaluations span A100 and H100, so their dispersion includes environment variation as well as decoding randomness; independent-training variation is measured by the separate 1.7B study.

\subsection{Further Analysis}

\begin{figure*}[t]
  \centering
  \includegraphics[width=0.98\textwidth]{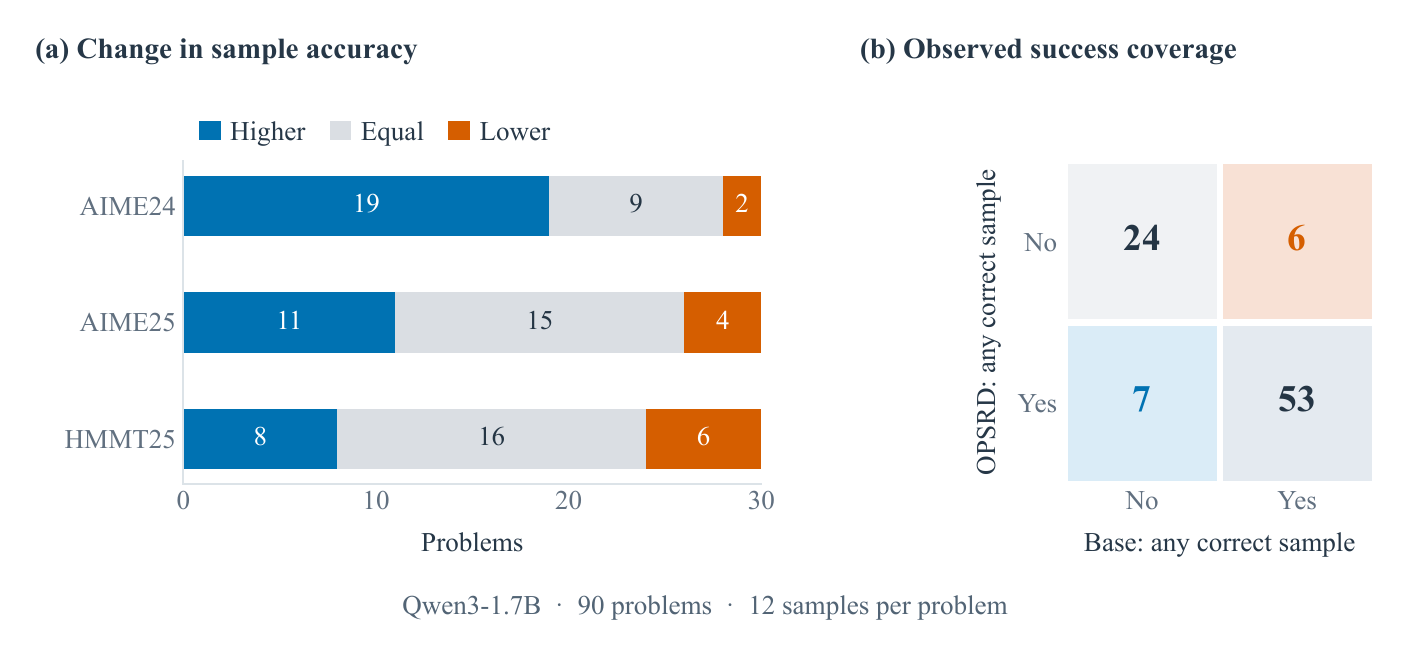}
  \caption{Problem-level changes at 1.7B. Left: problems with higher, equal, or lower sample accuracy than Base. Right: observed success coverage within 12 samples. The main gain is more reliable success on partially solved problems.}
  \label{fig:outcomes}
\end{figure*}

\paragraph{How the answers improve.}
An increase in average accuracy can reflect more reliable solutions to familiar problems or successful attempts on previously unsolved ones. We distinguish these outcomes by pairing Base and \method{} on every test problem. Figure~\ref{fig:outcomes} shows that \method{} increases the 12-sample success rate on 38 problems and lowers it on 12, with the rest tied. The improvements are most widespread on AIME 2024. This pattern accounts for the stronger AIME gains in the aggregate comparison and the repeated-training summary.

The change in observed coverage is much smaller. Seven problems with no correct Base sample receive a correct \method{} sample, while six change in the opposite direction. Most improved problems already have at least one successful Base trajectory. The main benefit is a better chance of reaching a correct answer on problems the student sometimes solves already. This is consistent with using a role teacher's local preferences to strengthen reasoning that the ordinary student can already express under its usual prompt.

This distinction connects the evaluation to the training design. \method{} learns on student-generated prefixes, so useful feedback must act within trajectories the student actually visits. The problem-level result is compatible with that local form of guidance. It also explains why an improvement in $\avg$ can coexist with little change in the number of problems solved at least once. Coverage here is measured at a finite sampling budget; broader claims about newly acquired problem-solving abilities would require a different evaluation.

\paragraph{Answer format and length.}
We also check whether the primary improvement can be explained by output failures. Format compliance exceeds 99\% across the primary conditions, and both Base and \method{} rarely reach the generation cap. The additional 68 correct generations exceed the differences in malformed and near-cap outputs. These discrepancies are too small individually to explain the gain, supporting the interpretation that the observed change concerns answer accuracy under the shared long-form evaluation budget.

\section{Discussion and Conclusion}

The experiments connect the role's placement to the distribution it teaches. Direct prompting changes the student's trajectory from its first token. Teacher-only distillation instead evaluates the trajectory the student already produced, and the placement controls favor keeping these two functions separate. Forward KL weights the update by the teacher's probabilities, allowing teacher-supported alternatives to receive a learning signal even when the student rarely predicts them. Equation~\ref{eq:fgrad} describes this behavior before clipping, while Appendix~\ref{app:algorithm} gives the gradient of the implemented objective. Tracking those alternatives through training would test how this local signal produces the observed answer-level gains. Both full and selective distillation improve Base across training seeds. Their paired interval crosses zero, so the selection result supports retaining the benefit at half the supervised positions. The implementation still requires both full forward passes.

\paragraph{Limitations.}
The evidence covers one model family, one mathematical training corpus, three competition benchmarks, and a matched 1.7B-to-8B scale suite. Three training seeds do not characterize every source of optimization variance. Evaluating other model families and tasks would test the broader reach of this training recipe.

\method{} turns a fixed expert role into supervision along the student's own reasoning. A frozen instance of the same base model provides the teacher distribution, and the student learns from it through forward KL at uncertain positions. Across the three tested scales, forward KL achieves the highest macro score among the evaluated divergences, with the clearest advantage over reverse KL at 1.7B. Training-seed repeats retain the gain over Base, and the trained student solves problems with an ordinary prompt. A role can therefore be useful even when direct prompting offers no measured improvement. Its teaching value lies in the guidance it supplies along the student's reasoning.

\section*{Reproducibility Statement}
Our code is publicly available at \url{https://github.com/zhansan114514/OPSRD}. The repository contains the training and evaluation code, the launch scripts with their hyperparameters, the analysis scripts used to summarize results and compute paired bootstrap intervals, tests for prompt isolation and the objective, and the recorded Python environment. Appendix~\ref{app:repro} lists the prompts and settings, Appendix~\ref{app:statistics} defines the resampling units, and Appendix~\ref{app:algorithm} specifies the loss reduction. Model checkpoints are not redistributed; the dataset and base models are public on Hugging Face.

\section*{AI Use Statement}
During the experiments, we used large language models (LLMs) as an auxiliary tool for experiment monitoring and management. Specifically, the LLMs were used to monitor experiment logs and runtime information, identify potential execution issues or anomalies, and assist in reporting the status of ongoing experiments. The LLMs did not determine the research questions, experimental methodology, hyperparameter settings, or final experimental conclusions. All experimental configurations, result verification, analysis, and scientific conclusions were determined and validated by the authors. We take full responsibility for the final content and results of this work.

\bibliography{references}
\bibliographystyle{iclr2027_conference}

\clearpage
\appendix
\section{Experimental Details}
\label{app:repro}

We provide the prompts, hyperparameters, and objective details needed to reproduce the comparisons. Our public repository (\url{https://github.com/zhansan114514/OPSRD}) contains the training and evaluation code, launch configurations, and scripts for summarizing results.

\subsection{Data and Hyperparameters}

Training uses the train split of \texttt{siyanzhao/Openthoughts\_math\_30k\_opsd}, containing 29,434 examples. Role distillation uses the problem field; Answer OPSD additionally provides the solution to its teacher. We reuse the published split without additional filtering or decontamination. The original runs did not pin a dataset revision. Table~\ref{tab:trainingconfig} summarizes training and evaluation, with the same final checkpoint used for all benchmarks.

\begin{table}[h]
\caption{Training and evaluation settings.}
\label{tab:trainingconfig}
\centering\small
\begin{tabular}{ll}
\toprule
Parameter & Setting \\
\midrule
Optimizer / steps & AdamW / 100 \\
Learning rate / maximum gradient norm & $5\times10^{-6}$ / 0.1 \\
Effective batch size / rollouts per problem & 36 / 1 \\
LoRA rank / scale & 64 / 128 \\
LoRA projections & q, k, v, o, gate, up, down \\
Training temperature / top-$p$ / top-$k$ & 1.1 / 0.95 / 20 \\
Training rollout cap & 1,024 tokens \\
Distillation temperature / vocabulary & 1.1 / Full \\
Contribution cap at 1.7B and 4B / at 8B & 0.05 / 0.06 \\
\rowcolor{methodblue}Selected fraction for \method{} & 0.50 per sequence \\
Precision / attention & BF16 / SDPA \\
\midrule
Evaluation samples per problem & 12 \\
Evaluation temperature / top-$p$ & 1.0 / 0.95 \\
Evaluation top-$k$ & Disabled \\
Evaluation generation cap / context limit & 38,912 / 40,960 tokens \\
\bottomrule
\end{tabular}
\end{table}

Answer OPSD is an internal reproduction with the same models, tasks, and metric as the cited study. Our effective batch of 36, SDPA attention, and fixed final checkpoint differ from its batch-32, FlashAttention-2 setup and checkpoint selection. Its published scores and our reproduced scores consequently refer to different training and evaluation configurations.

\subsection{Prompts}
\label{app:prompt}

\paragraph{Expert role.}
The teacher receives the following system message for every problem.
\begin{quote}\small
You are an expert olympiad mathematician. Solve the problem with rigorous, independent reasoning. Identify the key structure before calculating, verify each non-trivial step, reconsider doubtful branches, and check the final answer.
\end{quote}

\paragraph{Neutral context.}
The generic control shares the expert teacher's thinking mode and system-message placement, with the following instruction. It controls for an additional instruction, without matching the expert message's length or semantics.
\begin{quote}\small
You are an assistant. Respond to the user's request clearly and carefully. Follow the requested output format, keep the response relevant, and review the answer before finishing.
\end{quote}

\paragraph{Problem and answer format.}
The training user message contains \texttt{Problem:}, the problem, two newlines, and the instruction below. The Qwen3 chat template disables student thinking and enables teacher thinking. Evaluation enables thinking and omits the literal \texttt{Problem:} prefix; these choices remain fixed across the prompt and placement controls.
\begin{quote}\small
Please reason step by step, and put your final answer within \texttt{\textbackslash boxed\{\}}.
\end{quote}
Shared-role and student-only variants add the expert message before student sampling. Each checkpoint is evaluated both with and without that system message.

\paragraph{Reference-solution teacher.}
Answer OPSD encloses the solution between \texttt{=== Reference Solution Begin ===} and \texttt{=== Reference Solution End ===}. It then adds the following instruction and the boxed-answer request above.
\begin{quote}\small
After reading the reference solution above, make sure you truly understand the reasoning behind each step---do not copy or paraphrase it. Now, using your own words and independent reasoning, derive the same final answer to the problem above. Think step by step, explore different approaches, and don't be afraid to backtrack or reconsider if something \mbox{doesn't work out}:
\end{quote}
All teacher variants score the student's sampled completion after their respective contexts. Role and neutral distillation omit the reference solution from the model input.

\section{Algorithm Details}
\label{app:algorithm}

\begin{figure*}[t]
  \centering
  \includegraphics[width=0.99\textwidth]{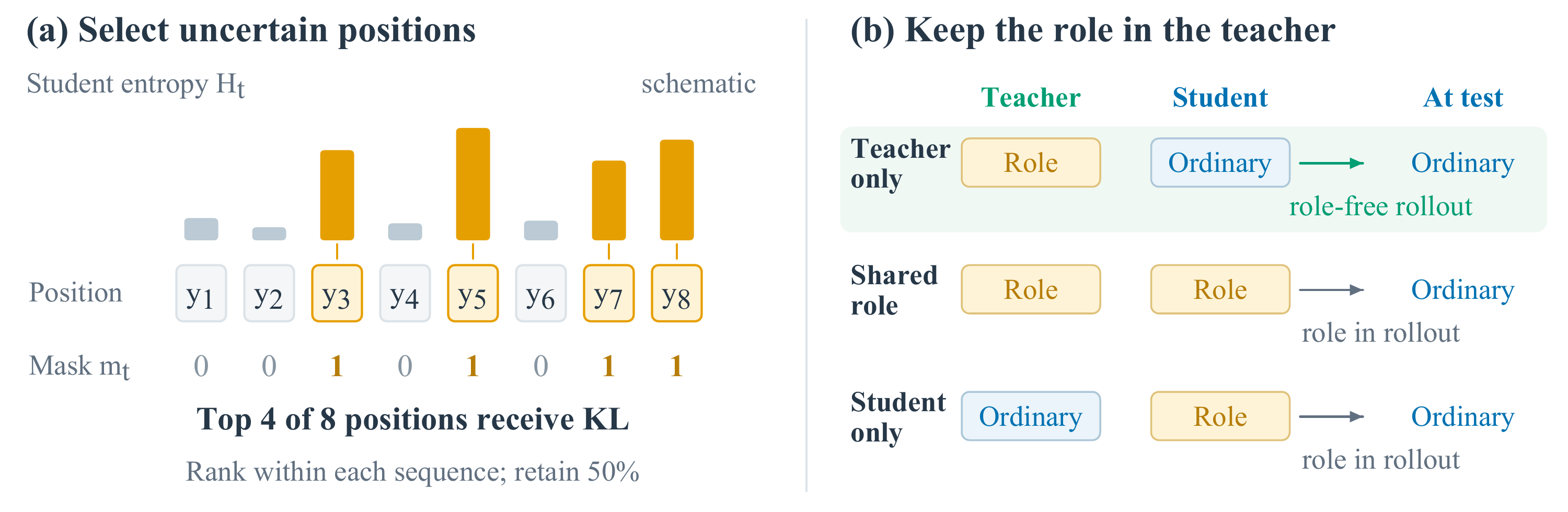}
  \caption{Selection and role placement. (a) Entropy ranking retains four of eight schematic positions. (b) Teacher-only conditioning keeps the student role-free during training and deployment.}
  \label{fig:methoddetails}
\end{figure*}

\subsection{Training Step}

For each batch, the adapted student first samples one completion per problem. The completion is appended to both contexts, and the student and frozen teacher score identical sampled positions. The teacher pass disables the student's adapter. Teacher probabilities and sampled tokens receive no gradients during student optimization.

Student entropy is computed from full-vocabulary float32 probabilities. Independently for each sequence, we retain the $\max(1,\lceil\rho T\rceil)$ valid positions with highest entropy (Figure~\ref{fig:methoddetails}). Padding and prompt positions are excluded. At the retained positions, we cap individual vocabulary contributions, sum over the full vocabulary, and average over the selected tokens. Entropy computation is chunked to reduce peak memory, preserving the same per-position values as the unchunked calculation.

The loss is evaluated per device and microbatch, after which distributed training averages local losses and accumulates gradients. With unequal selected-token counts, this reduction differs from weighting every token globally across devices. The implementation preserves the local reduction for all comparisons. Both full forward passes remain necessary after entropy selection.

\subsection{Clipped Gradient}

For a selected position, let $A_t=\{v:d_{t,v}<\tau\}$ denote contributions below the cap. Away from the cap boundary, the derivative of the implemented loss with respect to a student logit is
\begin{equation}
 \frac{\partial\ell_t}{\partial z_{S,j}^t}
 =\frac{1}{\eta}\left(p_S^t(j)\sum_{v\in A_t}p_T^t(v)
 -\mathbf{1}[j\in A_t]p_T^t(j)\right).
 \label{eq:clipgrad}
\end{equation}
Only contributions below the cap differentiate through $-p_T^t(v)\log p_S^t(v)$. When every contribution is active, the gradient reduces to $(p_S^t(j)-p_T^t(j))/\eta$. The framework selects a one-sided derivative at the cap boundary. Capped vocabulary entries have zero direct derivative but still affect other entries through softmax normalization. Teacher outputs and the entropy ranking remain detached.

\section{Additional Results}
\label{app:fullresults}

\subsection{Benchmark Scores}

Table~\ref{tab:fullscale} gives per-benchmark results for the primary evaluation at each model size. The 4B values here correspond to one evaluation of the same checkpoints used in the repeated-evaluation summary in the main paper. Each condition has 12 samples for each of the 90 test problems.

\begin{table*}[t]
\caption{Per-benchmark scores from the primary evaluation at each scale.}
\label{tab:fullscale}
\centering\small
\setlength{\tabcolsep}{4.5pt}
\begin{tabular}{llrrrrr}
\toprule
Model & Method & AIME24 & AIME25 & HMMT25 & Macro & $\Delta$ Base \\
\midrule
Qwen3-1.7B & Base & 46.67 & 36.11 & 24.44 & 35.74 & $0.00$ \\
Qwen3-1.7B & Answer OPSD & 51.39 & 39.17 & 25.56 & 38.70 & $+2.96$ \\
\rowcolor{methodblue}Qwen3-1.7B & \method{} & 57.78 & 41.67 & 26.67 & 42.04 & $+6.30$ \\
Qwen3-4B & Base & 73.61 & 66.11 & 41.11 & 60.28 & $0.00$ \\
Qwen3-4B & Answer OPSD & 75.00 & 68.61 & 43.89 & 62.50 & $+2.22$ \\
\rowcolor{methodblue}Qwen3-4B & \method{} & 72.22 & 67.78 & 44.44 & 61.48 & $+1.20$ \\
Qwen3-8B & Base & 75.56 & 69.44 & 43.33 & 62.78 & $0.00$ \\
Qwen3-8B & Answer OPSD & 75.83 & 71.67 & 45.28 & 64.26 & $+1.48$ \\
\rowcolor{methodblue}Qwen3-8B & \method{} & 77.22 & 71.39 & 48.89 & 65.83 & $+3.06$ \\
\bottomrule
\end{tabular}
\end{table*}

\subsection{Divergence Ablation}
\label{app:kl}

\begin{table*}[t]
\caption{Per-benchmark divergence results. JSD uses recorded aggregate scores.}
\label{tab:klfull}
\centering\small
\setlength{\tabcolsep}{4.5pt}
\begin{tabular}{lllrrrr}
\toprule
Model & Objective & GPU & AIME24 & AIME25 & HMMT25 & Macro \\
\midrule
\rowcolor{methodblue}1.7B & Forward KL & A100 & 57.78 & 41.67 & 26.67 & \textbf{42.04} \\
1.7B & Reverse KL & A100 & 50.56 & 35.56 & 25.56 & 37.22 \\
1.7B & JSD & H100 & 51.39 & 38.61 & 22.22 & 37.41 \\
\rowcolor{methodblue}4B & Forward KL & A100 & 72.22 & 67.78 & 44.44 & \textbf{61.48} \\
4B & Reverse KL & A100 & 74.72 & 67.50 & 40.56 & 60.93 \\
4B & JSD & H100 & 71.94 & 65.28 & 43.06 & 60.09 \\
\rowcolor{methodblue}8B & Forward KL & A100 & 77.22 & 71.39 & 48.89 & \textbf{65.83} \\
8B & Reverse KL & A100 & 75.00 & 70.00 & 46.39 & 63.80 \\
8B & JSD & H100 & 75.56 & 69.17 & 45.28 & 63.33 \\
\bottomrule
\end{tabular}
\end{table*}

Forward and reverse KL respectively use $\mathrm{KL}(p_T\|p_S)$ and $\mathrm{KL}(p_S\|p_T)$. JSD uses the equally weighted mixture $M=(p_T+p_S)/2$,
\begin{equation}
 D_{\mathrm{JSD}}=\tfrac12\mathrm{KL}(p_T\|M)+\tfrac12\mathrm{KL}(p_S\|M).
\end{equation}
Each objective retains the full vocabulary, applies the same per-contribution upper cap, and reduces over the selected positions. All variants use the expert teacher, role-free student, 50\% entropy fraction, and the training budget in Table~\ref{tab:trainingconfig}.

The KL runs use A100 and JSD uses H100. Device counts and microbatch layouts differ in some runs while preserving the nominal effective batch; local token reduction and rollout scheduling can therefore vary. The paired forward-versus-reverse intervals describe problem uncertainty for one training run per condition. They are $[2.41,7.31]$, $[-1.85,2.96]$, and $[-0.56,4.81]$ at 1.7B, 4B, and 8B. JSD scores are supported by contemporaneous aggregate records; per-sample JSD outputs are not publicly available.

\subsection{Role Placement}

Table~\ref{tab:fulllocus} separates the placement comparison by benchmark. The role-free student and role teacher define the teacher-only condition. Shared role conditions both models; student only conditions the rollout generator. All variants use the same 1.7B protocol and selection fraction. Relative to teacher-only training with ordinary evaluation, shared-role and student-only training have paired macro intervals of $[-5.83,-0.37]$ and $[-6.48,-1.30]$ under ordinary evaluation.

\begin{table*}[t]
\caption{Complete placement ablation, relative to teacher-only training with no test role.}
\label{tab:fulllocus}
\centering\small
\setlength{\tabcolsep}{4.5pt}
\begin{tabular}{llrrrrr}
\toprule
Training role & Test role & AIME24 & AIME25 & HMMT25 & Macro & $\Delta$ \\
\midrule
\rowcolor{methodblue}Teacher only & No role & 57.78 & 41.67 & 26.67 & 42.04 & $0.00$ \\
Teacher only & Expert role & 51.11 & 41.94 & 25.28 & 39.44 & $-2.59$ \\
Shared role & No role & 52.22 & 40.83 & 23.89 & 38.98 & $-3.06$ \\
Shared role & Expert role & 47.50 & 39.44 & 23.89 & 36.94 & $-5.09$ \\
Student only & No role & 50.28 & 39.72 & 24.72 & 38.24 & $-3.80$ \\
Student only & Expert role & 47.22 & 38.33 & 23.33 & 36.30 & $-5.74$ \\
\bottomrule
\end{tabular}
\end{table*}

\subsection{Selection Sensitivity}
\label{app:entropy}

Selected positions have much higher student entropy than omitted positions (Table~\ref{tab:entropydiag}). This confirms that the mask concentrates the objective on uncertain predictions. The measured divergence also shows a nonzero teacher signal at the retained positions.

\begin{table}[t]
\caption{Entropy selection in the primary run.}
\label{tab:entropydiag}
\centering\small
\setlength{\tabcolsep}{4.5pt}
\begin{tabular}{lr}
\toprule
Quantity & Mean \\
\midrule
Selected valid-token fraction & 0.500 \\
Selected-position entropy & 0.874 \\
Omitted-position entropy & 0.023 \\
Unclipped selected divergence & 0.194 \\
\bottomrule
\end{tabular}
\end{table}

The fraction pilot compares 25\%, 50\%, and 75\% selection with four samples per problem and the same generation cap as the main evaluation (Table~\ref{tab:fractionpilot}). The 50\% reference uses its first four matched samples. The 25\% result ties the reference and 75\% scores lower. Neither reaches the one-point improvement required for a full-budget follow-up. These exploratory measurements motivate retaining the default fraction; the principal density comparison uses the \mbox{complete 50\% and 100\% evaluations}.

\begin{table}[t]
\caption{Selection-fraction pilot with four samples per problem.}
\label{tab:fractionpilot}
\centering\small
\setlength{\tabcolsep}{4.5pt}
\begin{tabular}{rrr}
\toprule
Fraction & Avg@4 & $\Delta$ 50\% \\
\midrule
25\% & 43.06 & $0.00$ \\
\rowcolor{methodblue}50\% & 43.06 & $0.00$ \\
75\% & 38.89 & $-4.17$ \\
\bottomrule
\end{tabular}
\end{table}

\section{Statistical Analysis}
\label{app:statistics}

\paragraph{Metric.}
For benchmark $b$ with 30 problems and $S=12$ samples per problem, let $a_{j,s}$ denote answer correctness. We compute
\begin{equation}
 \mathrm{Avg@12}_b=\frac{100}{30S}\sum_{j=1}^{30}\sum_{s=1}^{S}a_{j,s}.
\end{equation}
The macro score averages the three benchmark scores. Equal problem and sample counts make it identical to pooled accuracy over 1,080 generations. Observed success coverage instead counts problems with any correct sample, which is the quantity contrasted with average accuracy in Figure~\ref{fig:outcomes}.

\paragraph{Paired intervals.}
We average each problem's 12 correctness labels, align conditions by benchmark and problem, and resample their paired accuracy differences. Per-task intervals resample 30 problems; pooled intervals resample all 90 problems. Primary comparisons use 10,000 bootstrap draws and the divergence comparison uses 50,000. This preserves the problem as the sampling unit rather than treating repeated generations as independent problems.

\paragraph{Training and evaluation repeats.}
For three-run training comparisons, each of 30,000 bootstrap draws resamples three runs and 30 problems within each benchmark. The same problem draw is shared across the sampled runs, preserving dependence from the common test set and fixed Base evaluation. Reported standard deviations are sample standard deviations across the stated repeats. The 4B repeats hold trained weights fixed and span A100 and H100; their dispersion combines decoding and environment variation. Training variability is assessed separately at 1.7B.

\paragraph{Study design.}
Intervals apply to the stated comparison without adjustment for multiple ablations. Additional training runs followed a favorable primary result, and alternative selection fractions were screened at a smaller sampling budget. These comparisons form an exploratory study. Each summary includes every completed repeat or model scale belonging to that comparison.

\section{Broader Impact}
\label{app:ethics}

The experiments use public mathematical problems and model-generated reasoning, with no human subjects or private records. Expertise labels require task-specific validation before deployment because a role can change model behavior unpredictably. Training uses student and teacher forward passes, while inference uses only the adapted student and its ordinary prompt.

\end{document}